\documentclass{article}

\usepackage[preprint,nonatbib]{neurips_2026}
\usepackage[numbers]{natbib}

\usepackage[utf8]{inputenc}
\usepackage[T1]{fontenc}
\usepackage{hyperref}
\usepackage{url}
\usepackage{booktabs}
\usepackage{amsfonts}
\usepackage{amsmath}
\usepackage{amssymb}
\usepackage{nicefrac}
\usepackage{microtype}
\usepackage{graphicx}
\usepackage{subcaption}
\usepackage{float}
\usepackage{xcolor}
\usepackage{titlesec}
\titlespacing*{\section}{0pt}{1.8ex plus 0.5ex minus 0.3ex}{1.0ex plus 0.2ex}
\titlespacing*{\subsection}{0pt}{1.2ex plus 0.4ex minus 0.2ex}{0.6ex plus 0.1ex}
\title{Neural Bayesian Anomaly Mitigation: A Robust Loss that Doubles as an Unsupervised Contamination Classifier}

\author{%
  S.~A.~K. Leeney \\
  Astrophysics Group, Cavendish Laboratory \\
  University of Cambridge \\
  \texttt{sakl2@cam.ac.uk} \\
  \And
  W.~J. Handley \\
  Institute of Astronomy \\
  University of Cambridge \\
  \texttt{wh260@cam.ac.uk} \\
  \And
  H.~T.~J. Bevins \\
  Astrophysics Group, Cavendish Laboratory \\
  University of Cambridge \\
  \texttt{htjb2@cam.ac.uk} \\
  \And
  E. de Lera Acedo \\
  Astrophysics Group, Cavendish Laboratory \\
  University of Cambridge \\
  \texttt{ed330@cam.ac.uk} \\
}

\begin{document}

\maketitle

\begin{abstract}
Engineered robust losses such as Huber, Student-$t$, and generalised cross-entropy make supervised models tolerant of contamination but cannot answer \emph{which} observations are corrupted. We introduce \emph{Neural Bayesian Anomaly Mitigation} (NBAM), a general-purpose drop-in loss derived from a Bayesian latent-switch mixture model: the marginal likelihood defines a robust supervised loss, and the associated posterior defines an unsupervised contamination classifier. Like Huber or Student-$t$, NBAM can replace the standard training loss in any supervised pipeline; unlike them, it additionally learns a structured contamination model and returns a calibrated per-sample contamination posterior. A learned input-dependent prior $\pi_\phi(x)$ captures the \emph{spatial locality} of contamination, so that samples near known corruptions are more likely to be flagged, while an Occam penalty emerges automatically and regularises against over-flagging. On CIFAR-10 with asymmetric label contamination, NBAM recovers the structure of the corruption process without supervision: the contamination posterior separates clean from corrupted samples, and the learned anomaly head identifies the direction of every label-flip pair. Alongside these capabilities, NBAM outperforms the four robust-loss baselines considered here at contamination rates $0.2$--$0.6$.
\end{abstract}

\begin{figure*}[t]
\centering
\includegraphics[width=0.97\textwidth]{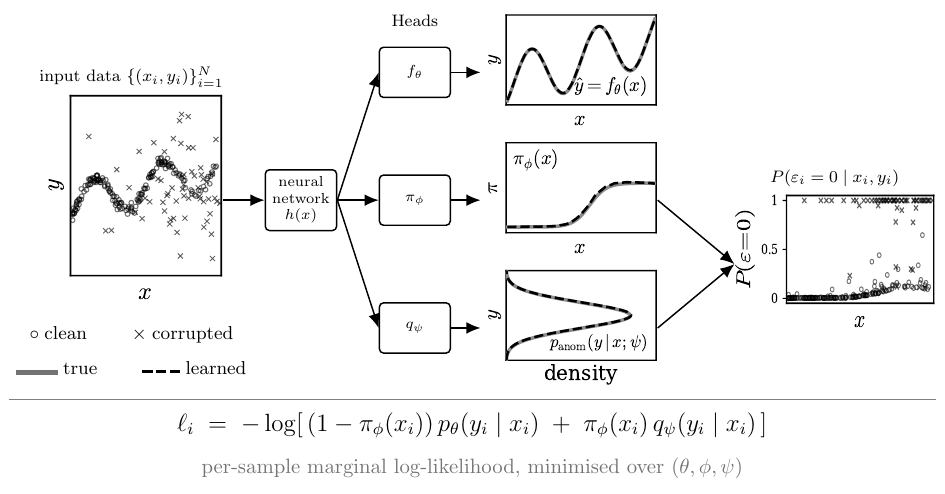}
\caption{NBAM architecture and synthetic demonstration. A shared neural network feeds three heads: the clean predictor $f_\theta$, the contamination prior $\pi_\phi(x)$, and the structured anomaly likelihood $q_\psi(y \mid x)$. Inset panels show each head's output on a 1D synthetic regression with contextual contamination (25\% mean rate, Gaussian corruption), alongside the resulting contamination posterior $P(\varepsilon_i = 0 \mid x_i, y_i)$, where $\varepsilon_i=0$ denotes contamination. All three heads closely recover their corresponding ground-truth functions from a single marginal-likelihood objective; the posterior separates clean from corrupted samples with 96.7\% accuracy. The loss below is the negative log marginal likelihood after analytically marginalising the latent contamination flag.}
\label{fig:architecture}
\end{figure*}

\section{Introduction}
\label{sec:intro}

Data contamination is pervasive in modern supervised learning, but it is rarely i.i.d.\ or structureless. In the label-contamination setting that we use for demonstration here, real annotation pipelines induce instance-dependent, structured corruption driven by ambiguity, annotator bias, and semantic similarity. The practically important failure mode is not arbitrary corruption, but \emph{plausible, spatially clustered} corruption: observed targets are often perturbed toward confusable alternatives, as in the asymmetric transition model of Patrini et al.\ \citep{patrini2017forward}, and contamination tends to concentrate in specific regions of the input space where the task is intrinsically ambiguous. Nor is the problem confined to annotation: learnt components are increasingly embedded in scientific measurement pipelines \citep{leeney2025radiometer}, where corrupted observations distort downstream inference unless they are modelled explicitly. Learning under contamination is therefore not only a problem of robustness but a problem of probabilistic explanation: for each sample, is the observed target better explained by the clean predictive model or by a contamination process?

The dominant response has been to engineer robust training objectives. The generalised cross-entropy (GCE) loss \citep{zhang2018gce} interpolates between mean absolute error and cross-entropy; Symmetric Cross Entropy adds a reverse-CE term \citep{wang2019symmetricce}; active-passive losses and normalised cross-entropy provide theoretical noise-tolerance guarantees \citep{ma2020apl}; Early-Learning Regularisation exploits the empirical observation that networks fit clean patterns before memorising noise \citep{liu2020elr}; and Active Negative Loss further refines the family \citep{ye2023active}. Classical Huber \citep{huber1964robust} and Student-$t$ losses reshape the per-sample objective so that suspicious observations exert less influence. These methods are valuable precisely because they make predictors more tolerant of contaminated supervision \citep{song2022survey}. Yet they are limited in a different respect: they return a robust predictor, but not a posterior probability that a particular sample has been contaminated.

Other method families address the problem from different directions. Loss-correction methods model the label-flip mechanism explicitly: Patrini et al.\ \citep{patrini2017forward} estimate a transition matrix between clean and corrupted labels, and bootstrapping approaches mix the current prediction into the training target \citep{reed2014bootstrapping}. Sample-selection methods such as Co-teaching \citep{han2018coteaching}, MentorNet \citep{jiang2018mentornet}, and DivideMix \citep{li2020dividemix} perform an implicit clean/noisy partition; DivideMix is closest to our setting, fitting a Gaussian mixture model to per-sample losses and then training with semi-supervised consistency regularisation. Probabilistic alternatives are closer in spirit: PENCIL treats labels as trainable distributions \citep{yi2019pencil}, Confident Learning estimates label errors from predictive probabilities \citep{northcutt2021confident}, joint optimisation frameworks alternate between updating model weights and updating estimated clean labels \citep{tanaka2018joint}, and recent multinomial-mixture approaches formulate explicit probabilistic models with identifiability guarantees \citep{nguyen2023multinomial}. Among these, Nguyen et al.\ \citep{nguyen2023multinomial} are the closest competitor: they also use a mixture structure to separate clean and corrupted supervision. However, these alternatives generally do not provide both a robust supervised objective and a deployable contamination classifier from the same training derivation.

We present \emph{Neural Bayesian Anomaly Mitigation} (NBAM), which closes this gap by returning both objects from one Bayesian derivation. NBAM is a general-purpose drop-in loss: it can replace the standard training objective in any supervised pipeline, just as Huber or Student-$t$ losses can, but additionally learns a structured contamination model. The construction extends the likelihood-level contamination model of Leeney et al.\ \citep{leeney2022rfi,leeney2025salt3} to supervised deep learning by introducing a latent binary contamination flag together with two learned heads: a prior $\pi_\phi(x)$ over whether an example is contaminated, and a structured anomaly likelihood $q_\psi(y \mid x)$ over how corrupted observations are distributed. Critically, $\pi_\phi(x)$ is a function of the input, so it captures the \emph{locality} of contamination: regions of the input space where corruption clusters receive higher prior weight, while clean regions are left undisturbed. Training minimises the negative log marginal likelihood after analytically marginalising the latent flag. Two consequences follow immediately. First, the marginal likelihood is a robust supervised loss: samples that are implausible under the clean predictor can be softly explained by the contamination branch rather than dominating the fit. Second, Bayes' rule yields a per-sample contamination posterior $P(\varepsilon_i = 0 \mid x_i, y_i)$, with $\varepsilon_i=0$ denoting contamination. Thus the trained network is simultaneously an unsupervised contamination classifier. These are not separate post-hoc mechanisms; they are the same posterior viewed from two angles.

A central feature of this construction is the Occam penalty: because the anomaly branch must be a proper normalised likelihood, assigning a sample to contamination incurs a probability-volume cost that discourages the degenerate ``flag everything'' solution. This penalty carries no tunable coefficient and acts as an automatic regulariser, giving NBAM a rate-agnostic training objective with no contamination-specific parameters to set.

On CIFAR-10 with Patrini-style asymmetric contamination \citep{patrini2017forward}, the trained model recovers the corruption process itself: the posterior $P(\varepsilon_i = 0 \mid x_i, y_i)$ separates clean from corrupted training samples, and the structured anomaly head identifies the flip direction of all four corrupted class pairs without ever seeing the contamination mask. None of the robust-loss baselines considered here returns any of these objects. Alongside these capabilities, NBAM outperforms standard cross-entropy, Student-$t$, Huber, and GCE \citep{zhang2018gce} at rates $0.2$--$0.6$, including a $1.2$-point gain over GCE at rate $0.4$.

Our contributions are fourfold:
\begin{enumerate}
\item We derive a general-purpose drop-in robust loss from the marginal likelihood of a Bayesian latent-switch mixture model, with no hand-added robustness terms. Like Huber or Student-$t$, NBAM can replace the standard loss in any supervised pipeline.
\item We obtain a per-sample contamination posterior $P(\varepsilon_i = 0 \mid x_i, y_i)$ with posterior-probability semantics, so the trained model doubles as an unsupervised contamination classifier; in the evaluated regimes, this posterior separates clean from corrupted samples.
\item We show that the learned heads recover the structure of the contamination process itself: the structured anomaly head $q_\psi(y \mid x)$ identifies the direction of the corruption mechanism, the input-dependent prior $\pi_\phi(x)$ captures the spatial locality of contamination clusters, and the emergent Occam penalty regularises over-flagging without requiring a tunable coefficient.
\item We demonstrate improved accuracy over the considered robust-loss baselines on CIFAR-10 under asymmetric contamination at rates $0.2$--$0.6$, together with competitive performance under symmetric contamination.
\end{enumerate}

Section~\ref{sec:theory} gives the Bayesian derivation of NBAM; Section~\ref{sec:method} specifies the experimental instantiation; and Section~\ref{sec:experiments} evaluates its empirical behaviour on synthetic and label-contamination benchmarks.

\section{Theory}
\label{sec:theory}

\subsection{Notation}
\label{sec:theory:bayes}

We write $\mathcal{L}(\theta) \equiv P(\mathcal{D} \mid \theta)$ for the likelihood, $p(\theta) \equiv P(\theta)$ for the parameter prior, and $\mathcal{Z} = \int \mathcal{L}(\theta)\,p(\theta)\,d\theta$ for the Bayesian evidence. The posterior is $P(\theta \mid \mathcal{D}) = \mathcal{L}(\theta)\,p(\theta) / \mathcal{Z}$. In what follows the only modification is the likelihood model entering~$\mathcal{L}$.

\subsection{Bayesian anomaly detection from first principles}
\label{sec:theory:bad}

We follow the likelihood-level anomaly-detection construction of Leeney et al.\ \citep{leeney2022rfi,leeney2025salt3}, which has been deployed for radio-frequency-interference excision and transient flagging in 21-cm cosmology \citep{leeney2022rfi,anstey2024reweighting} and for automated data curation in type-Ia supernova cosmology \citep{leeney2025salt3}, the latter built on differentiable SALT modelling \citep{leeney2025bandflux}. For conditionally independent data $\mathcal{D} = \{D_i\}_{i=1}^N$, the standard likelihood $P(\mathcal{D} \mid \theta) = \prod_i \mathcal{L}_i(\theta)$ is mis-specified when some observations are contaminated: a corrupted datum is not a low-probability draw from the clean model but a draw from a different process, and forcing it through $\mathcal{L}_i$ can distort inference on~$\theta$.

To represent this, introduce a latent binary mask $\varepsilon_i \in \{0,1\}$ ($\varepsilon_i\!=\!1$: clean, $\varepsilon_i\!=\!0$: contaminated) with per-sample likelihood $P(D_i \mid \theta, \varepsilon_i) = \mathcal{L}_i(\theta)^{\varepsilon_i}\,\Delta_i^{-(1-\varepsilon_i)}$, where $\Delta_i^{-1}$ is a broad uniform anomaly density \citep{leeney2022rfi,leeney2025salt3}. Assigning each $\varepsilon_i$ an independent Bernoulli prior with contamination probability~$\omega_i$ and multiplying gives the joint
\begin{equation}
P(\mathcal{D}, \varepsilon \mid \theta, \omega) = \prod_{i=1}^N \bigl[\mathcal{L}_i(\theta)(1-\omega_i)\bigr]^{\varepsilon_i} \bigl[\omega_i / \Delta_i\bigr]^{1-\varepsilon_i}.
\end{equation}
Because the factors are independent across~$i$, marginalising over all $2^N$ mask configurations separates exactly into the product
\begin{equation}
P(\mathcal{D} \mid \theta, \omega) = \prod_{i=1}^N \left[\mathcal{L}_i(\theta)(1-\omega_i) + \frac{\omega_i}{\Delta_i}\right].
\label{eq:exact-marginal-classical}
\end{equation}
This is differentiable everywhere and is the form we carry forward. Under the dominant-mask approximation, $\log P(\mathcal{D} \mid \theta, \omega) \approx \sum_i \max\!\bigl\{\log \mathcal{L}_i + \log(1\!-\!\omega_i),\;\log \omega_i - \log \Delta_i\bigr\}$, and the hard assignment rule becomes $\varepsilon_i^{\max}\!=\!1 \iff \log \mathcal{L}_i + \log \Delta_i > \operatorname{logit}(\omega_i)$. Thus the clean/contaminated classification emerges from Bayesian model comparison, with the familiar logit appearing naturally as the log-odds coordinate.

\paragraph{The Occam penalty.}  In the log likelihood, the anomaly contribution $\log \omega_i - \log \Delta_i$ contains the term $-\log \Delta_i$; equivalently, in the negative log objective it contributes the cost $+\log \Delta_i$. We call this probability-volume cost the \emph{Occam penalty}. It is not engineered; it appears because the anomaly branch must be a proper normalised density. The more diffuse that density, the less it can assign to any one observation. This cost helps prevent the trivial ``flag everything'' solution: without it, routing hard-to-fit data through the anomaly branch would be much less constrained. The Bayesian marginal likelihood supplies this constraint directly \citep{mackay2003information}.

\paragraph{Per-sample contamination posterior.}  Bayes' rule on the latent switch gives the model-implied posterior $P(\varepsilon_i\!=\!0 \mid D_i, \theta, \omega_i) = (\omega_i/\Delta_i)\,/\,[\mathcal{L}_i(\theta)(1\!-\!\omega_i) + \omega_i/\Delta_i]$. One derivation thus yields three objects: a contamination-aware marginal likelihood, an emergent Occam penalty, and a per-sample contamination posterior.

\subsection{Neural Bayesian anomaly detection}
\label{sec:theory:nbad}

In classical implementations, the contamination prior is specified rather than learned and $1/\Delta_i$ is uniform, so the anomaly model is static and structureless. \emph{Neural Bayesian Anomaly Mitigation (NBAM)} closes this gap by replacing both with learned distributions while leaving the Bayesian derivation unchanged.

We now specialise to supervised learning with $\mathcal{D} = \{(x_i, y_i)\}_{i=1}^N$ and clean predictive likelihood
\begin{equation}
\mathcal{L}_i(\theta) = p_\theta(y_i \mid x_i).
\end{equation}
NBAM replaces the static contamination prior and the static uniform contamination density by learned heads,
\begin{equation}
\pi_\phi(x_i) = \sigma(g_\phi(x_i)), \qquad q_\psi(y_i \mid x_i),
\end{equation}
where $\sigma(z) = (1 + e^{-z})^{-1}$ is the logistic sigmoid. Here $\pi_\phi(x_i) = P(\varepsilon_i = 0 \mid x_i, \phi)$ is the prior probability of contamination for input $x_i$. Because $\pi_\phi$ is a function of the input, it captures the \emph{spatial locality} of contamination: samples in regions of the input space where corruption clusters will receive higher prior weight, while samples in clean regions are left undisturbed. Two complementary components are learned: $\pi_\phi(x)$ encodes the \emph{locations} of contamination (where in input space corruption is likely), while $q_\psi(y \mid x)$ encodes its \emph{structure} (what corrupted outputs look like). Both are inferred jointly with the predictor.

The latent switch is then marginalised exactly as in the classical derivation, yielding a two-component mixture model:
\begin{equation}
\begin{aligned}
P(y_i \mid x_i, \theta, \phi, \psi)
&= \sum_{\varepsilon_i \in \{0,1\}} P(y_i \mid x_i, \theta, \psi, \varepsilon_i) \, P(\varepsilon_i \mid x_i, \phi) \\
&= (1 - \pi_\phi(x_i)) \, p_\theta(y_i \mid x_i) + \pi_\phi(x_i) \, q_\psi(y_i \mid x_i).
\end{aligned}
\end{equation}
Assuming conditional independence across samples,
\begin{equation}
P(\mathcal{D} \mid \theta, \phi, \psi) = \prod_{i=1}^N \left[(1 - \pi_\phi(x_i)) \, p_\theta(y_i \mid x_i) + \pi_\phi(x_i) \, q_\psi(y_i \mid x_i)\right].
\end{equation}
Training maximises this marginal likelihood, or equivalently minimises the negative log marginal likelihood,
\begin{equation}
\ell_i^{\mathrm{NBAM}} = -\log\!\left[(1 - \pi_\phi(x_i)) \, p_\theta(y_i \mid x_i) + \pi_\phi(x_i) \, q_\psi(y_i \mid x_i)\right], \qquad \mathcal{J}(\theta, \phi, \psi) = \sum_{i=1}^N \ell_i^{\mathrm{NBAM}}.
\label{eq:nbad-loss}
\end{equation}

The Occam penalty is preserved exactly because the training loss is the negative log marginal likelihood of this latent-switch mixture. Under the dominant-term approximation,
\begin{equation}
\log P(\mathcal{D} \mid \theta, \phi, \psi) \approx \sum_{i=1}^N \max\!\left\{\log p_\theta(y_i \mid x_i) + \log(1 - \pi_\phi(x_i)),\ \log \pi_\phi(x_i) + \log q_\psi(y_i \mid x_i)\right\}.
\end{equation}
If sample $i$ is routed through the anomaly branch, the negative log contribution is
\begin{equation}
-\log \pi_\phi(x_i) - \log q_\psi(y_i \mid x_i).
\end{equation}
This is the direct neural analogue of the classical penalty $-\log \omega_i + \log \Delta_i$. When $q_\psi(y_i \mid x_i) = 1/\Delta_i$, the second term reduces exactly to $+\log \Delta_i$; more generally, normalisation of $q_\psi$ enforces the same probability-volume penalty in learned, $x$-dependent form.

The Occam penalty therefore acts as an automatic regulariser against over-flagging: no regularisation coefficient is chosen and no sparsity term is added, yet every flag incurs a probability-volume cost determined by the normalised anomaly likelihood. Where standard robust losses such as Huber, Student-$t$, or GCE suppress outlier gradients by construction, NBAM obtains robustness through marginalisation over the latent contamination indicator together with the normalisation constraint on $q_\psi$.

\paragraph{No rate-specific tuning.}
Without a probability-volume penalty, a flexible contamination branch would be encouraged to absorb hard examples. The Occam penalty discourages that behaviour by attaching a marginal-likelihood cost to every flag. The learned prior $\pi_\phi(x)$ avoids specifying a global contamination rate, $q_\psi(y \mid x)$ avoids fixing a uniform or class-independent contamination distribution, and the posterior over $\varepsilon_i$ supplies a model-implied contamination probability. No contamination annotations are required.

\subsection{Robust supervised loss and per-sample contamination posterior}
\label{sec:theory:synthesis}

\paragraph{Per-sample contamination posterior.} Applying Bayes' rule to the same latent switch used in the training objective gives
\begin{equation}
P(\varepsilon_i = 0 \mid x_i, y_i, \theta, \phi, \psi) = \frac{\pi_\phi(x_i) \, q_\psi(y_i \mid x_i)}{(1 - \pi_\phi(x_i)) \, p_\theta(y_i \mid x_i) + \pi_\phi(x_i) \, q_\psi(y_i \mid x_i)}.
\label{eq:posterior}
\end{equation}
Because training never observes $\varepsilon_i$, NBAM simultaneously acts as an unsupervised contamination classifier: for each sample it returns the posterior probability that the observation has been generated by the contamination branch rather than the clean predictive model.

\paragraph{Robust supervised loss.} The per-sample loss $\ell_i^{\mathrm{NBAM}}$ from \eqref{eq:nbad-loss} is simultaneously a robust supervised objective: when the clean predictor explains $(x_i, y_i)$ well, the clean branch dominates and the loss approaches the usual negative log-likelihood; when the clean explanation is implausible, the anomaly branch absorbs the sample softly, reducing the influence of corrupted targets on the fit. Marginalising $\varepsilon$ yields the robust loss; conditioning on $\varepsilon$ yields the model-implied contamination posterior. These are the same Bayesian object viewed from two angles.

\section{Method}
\label{sec:method}

We instantiate the Bayesian mixture model of \S\ref{sec:theory:nbad} with small multilayer perceptrons and a fixed training protocol used unchanged across all contamination regimes. This is deliberate: NBAM is not given a condition-specific contamination-rate parameter, and the balance between clean and anomalous explanations is set by the marginal likelihood in \eqref{eq:nbad-loss} together with the Occam penalty described in \S\ref{sec:theory:nbad}, rather than by per-condition retuning.

\subsection{Synthetic demonstration}
\label{sec:method:synthetic}

To visualise the mechanism directly, we first study a 1D regression problem with contextual contamination. Clean targets follow $y = \sin(2.3x) + 0.3x$ for $x \in [-3,3]$, while the true contamination prior $\pi^\star(x)$ is a sigmoid ramp with mean rate $0.25$. Contaminated responses are drawn from $\mathcal{N}(0, 1.5^2)$, centred on the origin rather than on the clean signal. We train the same three-head architecture, now with width $48$, learning rate $10^{-2}$, and 20{,}000 optimisation steps, using a resample-each-step scheme to smooth the Monte Carlo objective. This experiment is purely illustrative: it shows, in a setting where the ground truth is known exactly, whether the clean predictor recovers the underlying signal, whether $\pi_\phi(x)$ recovers the contextual contamination pattern, and whether the learned anomaly head recovers the corruption distribution.

\subsection{Architecture}
\label{sec:method:architecture}

For an input feature vector $x$, the shared backbone $h(x)$ is a two-hidden-layer MLP with $\tanh$ activations,
\begin{equation}
h(x) = \tanh\!\bigl(W_2 \tanh(W_1 x + b_1) + b_2\bigr),
\end{equation}
with width $\texttt{hidden\_dim}$. Three heads branch from this shared representation. The clean predictor $f_\theta$ outputs $K$ logits for classification, defining $p_\theta(y \mid x)$ by a softmax; in the 1D regression demonstration the same head outputs a scalar predictive mean. The contamination-prior head outputs a scalar logit that is mapped to
\begin{equation}
\pi_\phi(x) = \sigma\!\bigl(g_\phi(h(x))\bigr),
\end{equation}
where $g_\phi$ is a small MLP and $\sigma$ is the logistic sigmoid. Because this head shares the learned representation $h(x)$, nearby inputs in feature space receive similar contamination priors, enabling $\pi_\phi$ to discover clusters of contamination. Finally, the structured anomaly head parameterises $q_\psi(y \mid x)$: for classification it outputs $K$ logits and hence a softmax distribution over classes; for regression it outputs $(\mu_a(x), \log \sigma_a^2(x))$, defining a Gaussian anomaly model.

For all CIFAR-10 experiments, the network operates on pre-extracted 512-dimensional ResNet-18 \citep{he2016resnet} ImageNet features rather than raw pixels. The backbone is therefore a lightweight MLP on fixed features.

\subsection{Training}
\label{sec:method:training}

Training minimises the NBAM negative log marginal likelihood from \eqref{eq:nbad-loss},
\begin{equation}
\ell_i = -\log\!\left[(1-\pi_\phi(x_i))\,p_\theta(y_i \mid x_i) + \pi_\phi(x_i)\,q_\psi(y_i \mid x_i)\right].
\end{equation}
For classification, $p_\theta(y \mid x)$ is the softmax categorical distribution induced by $f_\theta(x)$; for the synthetic regression demonstration, the clean branch is a Gaussian predictive model and the anomaly branch is the learned Gaussian head described above. We optimise all parameters jointly with Adam \citep{kingma2015adam} ($\beta_1=0.9$, $\beta_2=0.999$) for a fixed number of epochs, with no early stopping and no condition-specific tuning.

\paragraph{Default hyperparameters.}
Every CIFAR-10 condition and every method uses the same settings: learning rate $10^{-3}$, hidden width 256, $\pi$-head width 128, 200 epochs, batch size 512, Student-$t$ $\nu=3.0$, Huber $\delta=1.0$. No hyperparameter is tuned per contamination type or rate. The same model is used for symmetric and asymmetric contamination at all rates. NBAM does not receive the true contamination rate as input and does not tune a robustness coefficient to each condition; the marginal-likelihood tradeoff instead includes the Occam penalty. A one-axis-at-a-time sensitivity analysis over learning rate, hidden width, training epochs, and batch size (Appendix~\ref{app:sensitivity}) confirms that the qualitative conclusions for the symmetric setting are stable across a range of settings.

\subsection{Baselines}
\label{sec:method:baselines}

All baselines use the same backbone $h(x)$ and the same clean predictor head $f_\theta$, so the clean predictive capacity is matched across methods. We compare against four baselines, each using common literature defaults. \emph{Standard CE} minimises the usual cross-entropy. \emph{Student-$t$} applies the Student-$t$ robust loss with $\nu=3.0$ in the same per-sample loss space. \emph{Huber} applies a Huber clipping rule with $\delta=1.0$ to the per-sample cross-entropy. \emph{GCE} uses the generalised cross-entropy loss with $q=0.7$ \citep{zhang2018gce}. Our method, \emph{NBAM}, uses the full three-head architecture with the learned anomaly head $q_\psi(y \mid x)$. The sensitivity analysis in Appendix~\ref{app:sensitivity} shows that moderate perturbations to the training hyperparameters in the symmetric sweep do not alter the qualitative ranking.

\subsection{Datasets and contamination protocols}
\label{sec:method:data}

All image experiments are conducted on CIFAR-10 \citep{krizhevsky2009cifar}, represented by fixed 512-dimensional ResNet-18 features \citep{he2016resnet}. We use a fixed stratified $90/10$ split of the original training set to obtain 45{,}000 training examples; evaluation uses the standard 10{,}000-image clean test set. The classification problem has $K=10$ classes throughout. We study two contamination regimes. In the \emph{symmetric} protocol, each training label is independently replaced with probability $\eta$ by a class drawn uniformly from all $K$ classes (including the correct one), with $\eta \in \{0.1, 0.2, 0.3, 0.4, 0.5, 0.6\}$. In the \emph{asymmetric} protocol of Patrini et al.\ \citep{patrini2017forward}, visually similar classes are pair-flipped: truck$\rightarrow$automobile, bird$\rightarrow$airplane, cat$\rightarrow$dog, and deer$\rightarrow$horse, with rates $\eta \in \{0.1, 0.2, 0.3, 0.4, 0.5, 0.6\}$. We stop at $\eta=0.6$ because beyond this rate the clean signal becomes too weak for any method to fit reliably under these training conditions; all methods degrade toward chance and meaningful comparison is no longer possible. The model, optimiser, and hyperparameters are identical across both regimes and all rates; only the contamination process changes.

\subsection{Evaluation}
\label{sec:method:evaluation}

The headline metric is clean-test accuracy (top-1 on the 10{,}000-image clean test set, averaged over five seeds). As a secondary diagnostic, we evaluate the contamination posterior from \eqref{eq:posterior} on the training set, where the contamination indicator is known by construction, by examining the separation between truly-clean and truly-contaminated samples. A one-axis-at-a-time sensitivity study appears in Appendix~\ref{app:sensitivity}.

\section{Experiments}
\label{sec:experiments}

We evaluate NBAM on a synthetic regression problem and two CIFAR-10 contamination regimes: symmetric and asymmetric pair-flip. All results use the fixed hyperparameters of \S\ref{sec:method} with no per-condition tuning.

\subsection{Synthetic demonstration}
\label{sec:experiments:synthetic}

Figure~\ref{fig:architecture} shows both the NBAM architecture and a controlled synthetic experiment in which the ground truth is known exactly. From a single marginal-likelihood objective, the clean predictor $f_\theta$ recovers the underlying signal despite 25\% contamination, the prior head $\pi_\phi(x)$ tracks the true sigmoid contamination ramp, and the structured anomaly head $q_\psi(y \mid x)$ approximates the true Gaussian corruption law. The resulting posterior $P(\varepsilon_i = 0 \mid x_i, y_i)$ cleanly separates clean from corrupted samples, achieving 96.7\% classification accuracy. This experiment isolates the core mechanism of NBAM: all three heads are learned jointly by the same marginal-likelihood objective.

\begin{table}[t]
\caption{Clean-test accuracy on CIFAR-10 (mean $\pm$ SEM, 5 seeds). Top: symmetric contamination. Bottom: asymmetric (Patrini pair-flip) contamination. Bold indicates the best method or methods per row.}
\label{tab:results}
\centering\small
\textbf{Symmetric contamination}\\[3pt]
\begin{tabular}{lccccc}
\toprule
$\eta$ & Std.~CE & Student-$t$ & Huber & GCE & NBAM \\
\midrule
0.1 & .867{\scriptsize$\pm$.001} & \textbf{.870{\scriptsize$\pm$.001}} & \textbf{.870{\scriptsize$\pm$.001}} & .859{\scriptsize$\pm$.001} & .867{\scriptsize$\pm$.001} \\
0.2 & .862{\scriptsize$\pm$.001} & \textbf{.868{\scriptsize$\pm$.001}} & .866{\scriptsize$\pm$.001} & .856{\scriptsize$\pm$.001} & .858{\scriptsize$\pm$.001} \\
0.3 & .856{\scriptsize$\pm$.001} & \textbf{.863{\scriptsize$\pm$.001}} & .860{\scriptsize$\pm$.001} & .856{\scriptsize$\pm$.001} & .856{\scriptsize$\pm$.001} \\
0.4 & .848{\scriptsize$\pm$.002} & \textbf{.857{\scriptsize$\pm$.000}} & .855{\scriptsize$\pm$.000} & .853{\scriptsize$\pm$.002} & .852{\scriptsize$\pm$.002} \\
0.5 & .831{\scriptsize$\pm$.001} & .843{\scriptsize$\pm$.002} & .840{\scriptsize$\pm$.002} & \textbf{.851{\scriptsize$\pm$.001}} & .847{\scriptsize$\pm$.002} \\
0.6 & .812{\scriptsize$\pm$.002} & .830{\scriptsize$\pm$.002} & .821{\scriptsize$\pm$.002} & \textbf{.843{\scriptsize$\pm$.002}} & .834{\scriptsize$\pm$.002} \\
\bottomrule
\end{tabular}\\[10pt]
\textbf{Asymmetric (pair-flip) contamination}\\[3pt]
\begin{tabular}{lccccc}
\toprule
$\eta$ & Std.~CE & Student-$t$ & Huber & GCE & NBAM \\
\midrule
0.1 & .865{\scriptsize$\pm$.001} & \textbf{.868{\scriptsize$\pm$.001}} & \textbf{.868{\scriptsize$\pm$.000}} & .858{\scriptsize$\pm$.001} & \textbf{.868{\scriptsize$\pm$.001}} \\
0.2 & .855{\scriptsize$\pm$.001} & .859{\scriptsize$\pm$.001} & .861{\scriptsize$\pm$.000} & .853{\scriptsize$\pm$.001} & \textbf{.862{\scriptsize$\pm$.001}} \\
0.3 & .830{\scriptsize$\pm$.002} & .843{\scriptsize$\pm$.002} & .844{\scriptsize$\pm$.003} & .846{\scriptsize$\pm$.001} & \textbf{.851{\scriptsize$\pm$.003}} \\
0.4 & .784{\scriptsize$\pm$.006} & .793{\scriptsize$\pm$.007} & .787{\scriptsize$\pm$.012} & .802{\scriptsize$\pm$.007} & \textbf{.814{\scriptsize$\pm$.005}} \\
0.5 & .710{\scriptsize$\pm$.011} & .699{\scriptsize$\pm$.008} & .703{\scriptsize$\pm$.010} & .677{\scriptsize$\pm$.008} & \textbf{.726{\scriptsize$\pm$.015}} \\
0.6 & .636{\scriptsize$\pm$.011} & .625{\scriptsize$\pm$.008} & .608{\scriptsize$\pm$.009} & .587{\scriptsize$\pm$.005} & \textbf{.642{\scriptsize$\pm$.010}} \\
\bottomrule
\end{tabular}
\end{table}

\subsection{CIFAR-10: Asymmetric contamination}
\label{sec:experiments:asym}

\begin{figure}[t]
\centering
\begin{subfigure}[b]{0.53\linewidth}
\centering
\includegraphics[width=\linewidth]{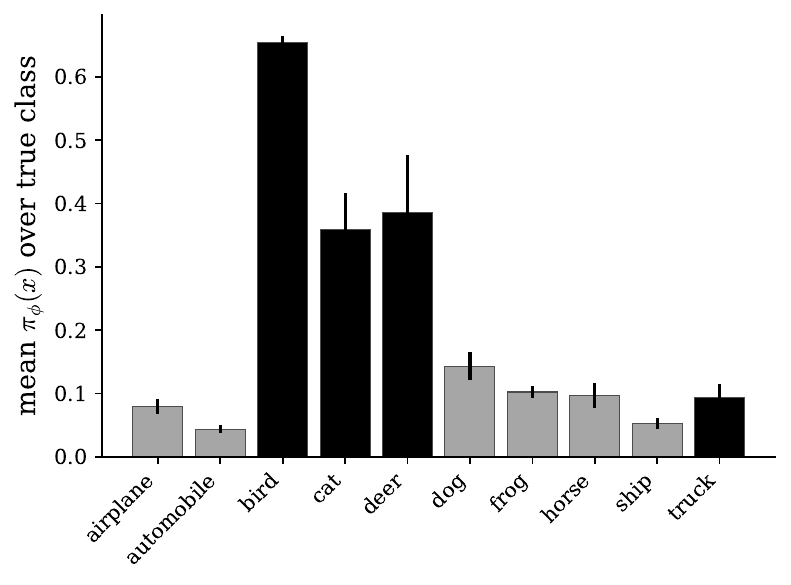}
\caption{learned contamination prior $\pi_\phi$}
\label{fig:pi-structure}
\end{subfigure}\hfill
\begin{subfigure}[b]{0.45\linewidth}
\centering
\includegraphics[width=\linewidth]{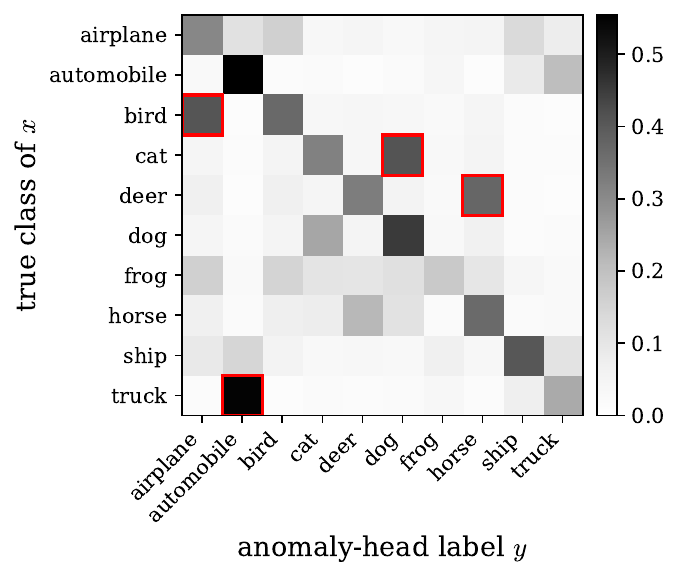}
\caption{mean $q_\psi(y \mid x)$ by true class}
\label{fig:q-structure}
\end{subfigure}
\caption{What the learned heads discover under asymmetric contamination ($\eta=0.4$, pooled over five seeds). (a)~Mean contamination prior $\pi_\phi(x)$ grouped by true class; dark bars mark the four pair-flip source classes (true flip rate $0.4$). The prior is elevated for bird, cat, and deer, but does not flag truck. (b)~Mean anomaly-head distribution $q_\psi(y \mid x)$ grouped by the true class of $x$; red boxes mark the expected flip targets. For all four source classes the modal anomaly-head label is exactly the flip target, recovering the directional corruption mechanism without supervision.}
\label{fig:pi-q-structure}
\end{figure}

The bottom panel of Table~\ref{tab:results} presents the asymmetric contamination results. Under the pair-flip protocol of Patrini et al.\ \citep{patrini2017forward}, NBAM outperforms every considered baseline at rates $0.2$--$0.6$ and ties for best at $0.1$. At $\eta=0.4$, NBAM reaches $0.814$ versus $0.802$ for GCE, a gain of $1.2$ accuracy points. This advantage arises from two sources, visualised directly in Figure~\ref{fig:pi-q-structure}. First, the structured anomaly head $q_\psi(y \mid x)$ learns the contamination mechanism itself: grouped by the true class of $x$, the modal anomaly-head label is exactly the flip target for all four corrupted classes (truck$\rightarrow$automobile, bird$\rightarrow$airplane, cat$\rightarrow$dog, deer$\rightarrow$horse; Figure~\ref{fig:q-structure}), so corruption is modelled as directional rather than class-independent. Second, the learned prior $\pi_\phi(x)$ tends to discover where contamination clusters in feature space: it is strongly elevated for three of the four flipped source classes (bird, cat, deer; near the true flip rate of $0.4$ for the latter two) while remaining near zero for uncorrupted classes (Figure~\ref{fig:pi-structure}). The truck class is the exception: its prior stays low, the truck$\rightarrow$automobile flips evidently absorbed by the clean branch instead, indicating that the locality captured by $\pi_\phi$ is a tendency rather than a guarantee. Since the per-sample posterior combines both heads with the clean-branch likelihood ratio, detection does not rest on $\pi_\phi$ alone (Figure~\ref{fig:posterior-separation}). This advantage is obtained with the same fixed hyperparameters at every rate, with the Occam penalty acting as an automatic regulariser against over-flagging.

\subsection{CIFAR-10: Symmetric contamination}
\label{sec:experiments:sym}

The top panel of Table~\ref{tab:results} shows the contrasting symmetric regime. At contamination rates $0.2$--$0.4$, all methods lie within roughly one accuracy point, consistent with i.i.d.\ uniform corruption being a setting in which these robust losses are already well matched to the problem. At $\eta=0.6$, NBAM outperforms standard cross-entropy, Student-$t$, and Huber, and trails only GCE. This pattern is consistent with the model intuition: when the corruption process is itself uniform over classes, the structured anomaly head has less structure to discover and provides less advantage over the baselines.

\subsection{Contamination posterior and accuracy curves}
\label{sec:experiments:posterior}

\begin{figure}[t]
\centering
\includegraphics[width=\linewidth]{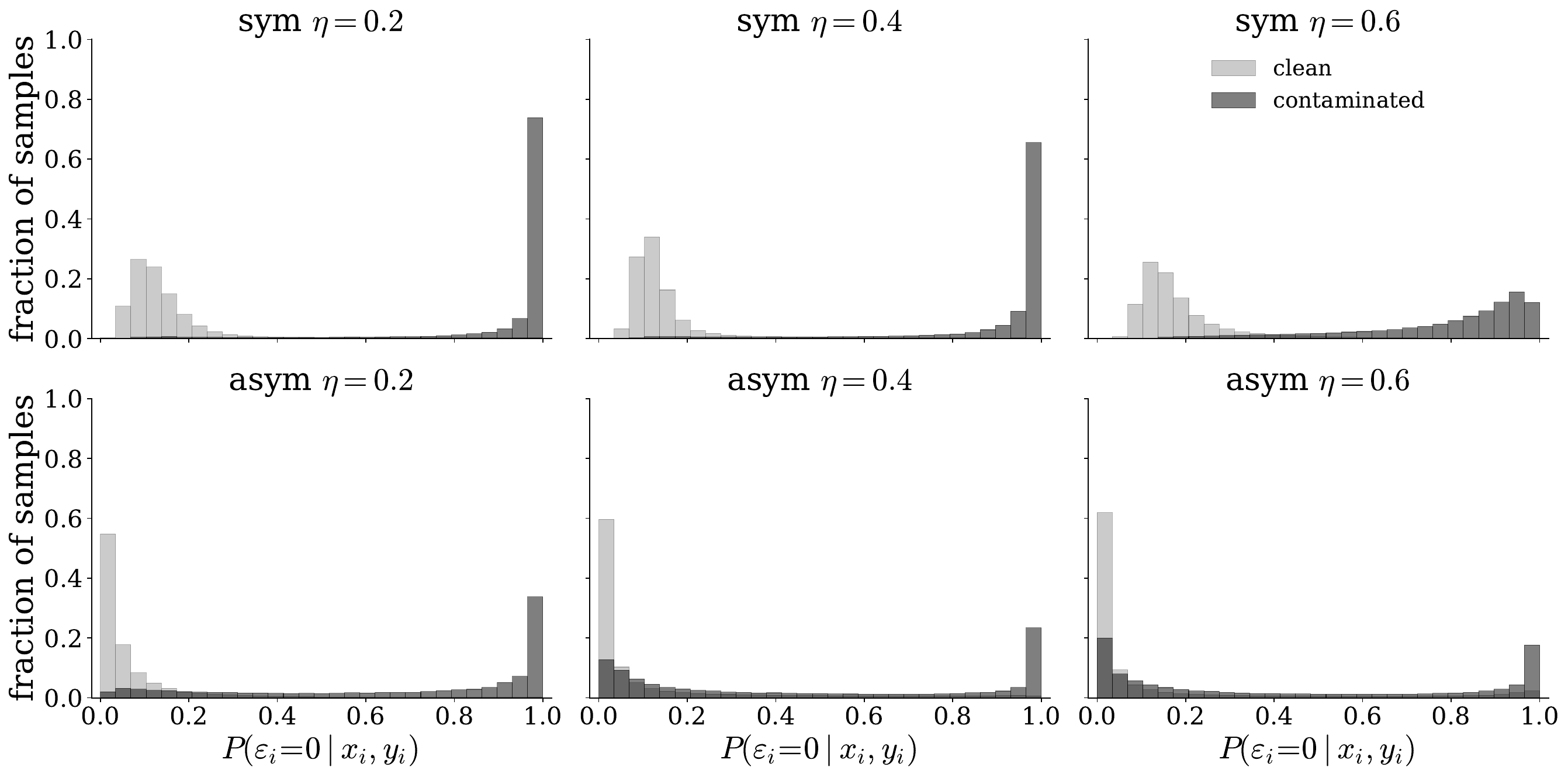}
\caption{Separation of the NBAM contamination posterior $P(\varepsilon_i\!=\!0 \mid x_i, y_i)$ on CIFAR-10 training data. Top row: symmetric contamination; bottom row: asymmetric pair-flip contamination, both at $\eta \in \{0.2, 0.4, 0.6\}$. Each panel overlays density-normalised histograms for truly-clean (light) and truly-contaminated (dark) samples, pooled over five seeds. Clean samples cluster near zero; contaminated samples cluster near one. Separation remains strong even at symmetric $\eta = 0.6$; at asymmetric $\eta = 0.6$ a fraction of the contaminated mass is assigned to the clean branch, as expected once pair-flips exceed the $0.5$ rate at which the corrupted label becomes the majority within each affected pair.}
\label{fig:posterior-separation}
\end{figure}

\begin{figure}[t]
\centering
\includegraphics[width=0.85\textwidth]{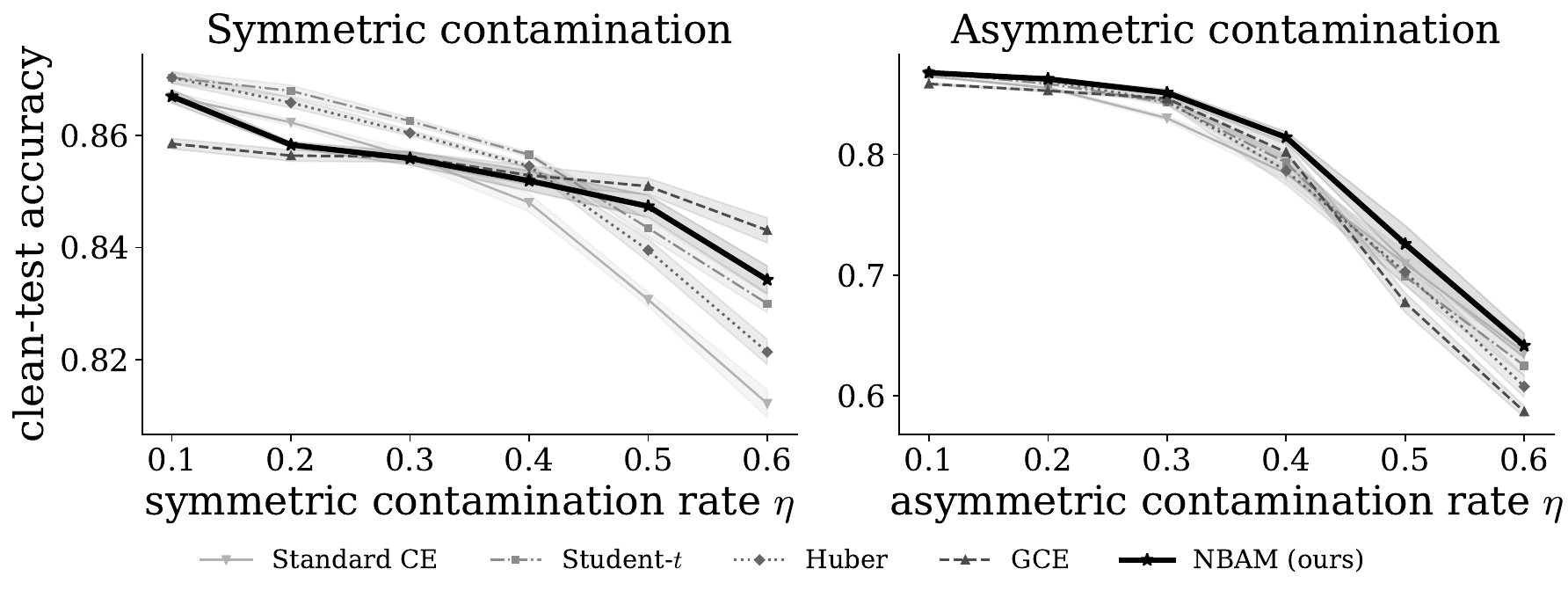}
\caption{Clean-test accuracy versus contamination rate for symmetric (left) and asymmetric pair-flip (right) contamination. NBAM ties for best at $\eta=0.1$ and leads at $\eta=0.2$--$0.6$ under asymmetric contamination.}
\label{fig:accuracy_vs_noise}
\end{figure}

Figure~\ref{fig:posterior-separation} evaluates the contamination posterior. Unlike loss-based anomaly scores, which require a chosen threshold or labelled validation data to make binary decisions, NBAM returns a posterior probability $P(\varepsilon_i\!=\!0 \mid x_i, y_i)$ with proper posterior-probability semantics. The histograms show clean separation between truly-clean and truly-contaminated training samples across both symmetric and asymmetric contamination regimes, confirming that the posterior is discriminative. Among the robust-loss baselines considered here, no method provides this object.

Notably, the posterior remains well-separated even at symmetric $\eta=0.6$, where 60\% of training labels are uniformly corrupted and the corruption carries no spatial or structural signal. This suggests that the posterior quality is driven primarily by the per-sample likelihood ratio between the clean and anomaly branches, rather than by the learned structure in $q_\psi$ or the locality captured by $\pi_\phi$. Under asymmetric contamination the separation is comparably sharp at $\eta \le 0.4$, while at $\eta = 0.6$ a fraction of contaminated samples receives low contamination posterior: once the pair-flip rate exceeds $0.5$, the corrupted label is the majority label within each affected pair, so a subset of flips is no longer identifiable from the data alone. The posterior degrades gracefully in this regime rather than collapsing, and the corresponding accuracy advantage of NBAM at asymmetric $\eta = 0.6$ persists (Table~\ref{tab:results}).

Figure~\ref{fig:accuracy_vs_noise} summarises the regime dependence: NBAM leads in the asymmetric pair-flip setting and remains competitive when contamination is uniform, with the same fixed hyperparameters throughout.

\section{Discussion}
\label{sec:discussion}

NBAM is not a specialised architecture for a particular contamination setting; it is a general-purpose training objective that can replace the standard loss in any supervised pipeline, exactly as Huber, Student-$t$, or GCE losses can. The key difference is that NBAM wraps the supervised loss inside a Bayesian mixture model rather than reshaping the per-sample gradient. The same NBAM loss applies to classification, regression, or any setting where a per-sample negative log-likelihood is available, with no modification to the base model or the optimiser.

The emergent Occam penalty is the central mechanism that makes NBAM work without contamination-rate tuning. Because the anomaly branch $q_\psi(y \mid x)$ must be a proper normalised density, routing a sample through it incurs a probability-volume cost that acts as an automatic regulariser against over-flagging. In a standard robust loss such as Huber, the degree of robustness is set by a fixed parameter ($\delta$); in NBAM, it is set by the marginal likelihood itself. The Occam penalty ensures that the model prefers the simpler explanation, that a sample is clean, unless the contamination branch can offer a substantially better fit. This is a direct consequence of Bayesian model comparison and requires no tunable coefficient.

A distinctive feature of NBAM is that the learned prior $\pi_\phi(x)$ is a function of the input, so the model can discover that contamination is \emph{spatially clustered}. In the asymmetric Patrini regime, cats and dogs share a similar feature representation, and the label flips cat$\rightarrow$dog concentrate in this region; $\pi_\phi(x)$ learns to assign higher contamination prior to inputs in that neighbourhood, while leaving unambiguous classes like trucks largely undisturbed. This locality is not available to standard robust losses, which treat each sample independently regardless of its position in input space. The combination of locality in $\pi_\phi(x)$ (the prior head learns \emph{where} contamination occurs) and structure in $q_\psi(y \mid x)$ (the anomaly head learns \emph{what} it looks like) enables NBAM to outperform baselines in the asymmetric regime, where contamination has both spatial and structural regularity. In the symmetric regime, where corruption is i.i.d.\ uniform over classes, there is no spatial structure for $\pi_\phi(x)$ to exploit, so the advantage of NBAM's learned heads is reduced; GCE is slightly better at the most extreme symmetric rates, consistent with its fixed $q$ parameter being well matched to uniform corruption.

Beyond robust prediction, NBAM produces a discriminative contamination posterior $P(\varepsilon_i = 0 \mid x_i, y_i)$ with no extra model. No classical robust loss provides this object. The posterior histograms in Figure~\ref{fig:posterior-separation} confirm clean separation across a range of conditions. This posterior could be used for data cleaning, active relabelling, or curriculum design. In practice, a practitioner could inspect the highest-posterior samples to diagnose systematic annotation failures, or feed the posterior into a downstream selection mechanism without training a separate anomaly detector, for instance into Bayesian experimental-design machinery \citep{leeney2026cnbre} to decide which samples are worth re-labelling.

The current formulation has limitations: the regression anomaly model is a single Gaussian, the structured head is less helpful under extreme symmetric corruption, and our experiments operate on pre-extracted features rather than end-to-end raw images. A broader evaluation including sample-selection and loss-correction methods would strengthen the empirical case. A natural direction for future work is \emph{composing} NBAM with existing robust losses: rather than wrapping a standard cross-entropy in the mixture model, one could wrap a Huber or GCE loss as the clean branch, potentially combining gradient-shaping benefits with structured contamination modelling. Other directions include richer anomaly parameterisations, end-to-end training on raw images, and extensions to domains beyond vision where contamination arises from measurement error, sensor faults, or systematic biases. Because NBAM requires only a per-sample negative log-likelihood, any such extension inherits the Occam penalty and contamination posterior without additional derivation.

\section{Conclusion}
\label{sec:conclusion}

NBAM shows that one Bayesian derivation can deliver both a competitive robust supervised loss and an unsupervised contamination classifier. By marginalising a latent contamination variable, the method returns a discriminative per-sample posterior and recovers the structure of the corruption process itself, identifying where contamination concentrates in input space and the direction of the label flips, while improving over the considered robust-loss baselines on asymmetric contamination at rates $0.2$--$0.6$ and remaining competitive under symmetric contamination. The key mechanisms are the Occam penalty, which regularises over-flagging via marginal-likelihood maximisation, and the input-dependent prior $\pi_\phi(x)$, which captures contamination locality. NBAM requires no contamination-rate tuning; everything follows from the Bayesian mixture model.

\bibliographystyle{abbrvnat}
\bibliography{references}

\clearpage
\appendix

\section{Technical appendix}
\label{app:technical}

\subsection{Sensitivity analysis}
\label{app:sensitivity}

We ran a one-axis-at-a-time sweep on the symmetric-contamination setting for $\eta \in \{0.1,0.2,\ldots,0.9\}$, comparing Standard CE, Student-$t$, Huber, GCE, and NBAM. Each sweep cell used five random seeds, with all non-swept hyperparameters fixed at the defaults in Table~\ref{tab:appendix-sweep}. Across all four axes, the symmetric-regime conclusions from the top panel of Table~\ref{tab:results} and the left panel of Figure~\ref{fig:accuracy_vs_noise} were unchanged: moderate perturbations in optimisation scale, model width, training length, or batch size shifted absolute accuracies only slightly and did not alter the qualitative regime structure. In particular, the crossover at which NBAM begins to outperform Standard CE, Student-$t$, and Huber remained in the same high-contamination regime for every hyperparameter setting. GCE likewise remained the strongest engineered baseline at the most extreme symmetric rates. This stability is consistent with the role of the Occam penalty: because routing a sample through the anomaly branch always incurs a marginal-likelihood cost, the model does not become systematically more aggressive under larger widths or longer training, and the crossover boundary is preserved without condition-specific retuning.

\begin{table}[t]
\caption{One-axis-at-a-time sensitivity sweep used for Appendix~\ref{app:sensitivity}. The reported main-text configuration is the default setting on each axis.}
\label{tab:appendix-sweep}
\centering\small
\begin{tabular}{lcc}
\toprule
Axis & Values & Default \\
\midrule
Learning rate & $\{10^{-4}, 3 \times 10^{-4}, 10^{-3}, 3 \times 10^{-3}\}$ & $10^{-3}$ \\
Hidden dimension & $\{64, 128, 256, 512\}$ & $256$ \\
Training epochs & $\{50, 100, 200, 400\}$ & $200$ \\
Batch size & $\{128, 256, 512, 1024\}$ & $512$ \\
\bottomrule
\end{tabular}
\end{table}

\subsection{Additional posterior-diagnostic details}
\label{app:posterior-diagnostic-details}

For the posterior separation histograms in Figure~\ref{fig:posterior-separation}, we evaluate the \emph{contamination} posterior rather than a thresholded loss score. With clean loss $\ell_i^{\mathrm{clean}} = -\log p_\theta(y_i \mid x_i)$, anomaly log-likelihood $a_i = \log q_\psi(y_i \mid x_i)$, and $\pi_i \equiv \pi_\phi(x_i)$, the posterior is
\begin{equation}
\hat{p}_i \equiv P(\varepsilon_i = 0 \mid x_i, y_i)
= \sigma\!\left(\operatorname{logit}(\pi_i) + \ell_i^{\mathrm{clean}} + a_i\right),
\end{equation}
which is the scalar form of Eq.~\eqref{eq:posterior}. Each histogram panel overlays density-normalised distributions of $\hat{p}_i$ for truly-clean ($z_i = 0$) and truly-contaminated ($z_i = 1$) training samples, pooled over five random seeds. Clean samples cluster near $\hat{p}_i \approx 0$ and contaminated samples near $\hat{p}_i \approx 1$, confirming that the posterior is discriminative even at high contamination rates.

\subsection{Reproducibility details}
\label{app:reproducibility}

All experiments were implemented in JAX and run on the C3 cluster using NVIDIA A100 GPUs. The CIFAR-10 image experiments used pre-extracted, frozen ResNet-18 ImageNet features; no feature extractor was fine-tuned. Unless noted otherwise, each reported condition used five random seeds and is summarised as mean $\pm$ standard error of the mean. The full symmetric-contamination sensitivity sweep of Appendix~\ref{app:sensitivity} required approximately two GPU-hours on a single A100. We will release the training code, configuration files, and plotting scripts to support reproduction of the results reported in the main text and appendix.

\begin{ack}
GPU compute for this work was provided by cthree.cloud (referred to as C3 in
the text), whose support we gratefully acknowledge. SL is funded by the ERC
through the UKRI guarantee scheme.
\end{ack}

\end{document}